\documentclass[letterpaper,10pt,conference]{ieeeconf}
\IEEEoverridecommandlockouts
\usepackage[T1]{fontenc}
\usepackage{amsmath,amssymb}
\usepackage{graphicx}
\usepackage{booktabs,array}
\usepackage{xcolor}
\usepackage{cite}
\usepackage[hidelinks]{hyperref}

\graphicspath{{images_updated_arXiv/}}
\title{\LARGE\bf Learning and Transferring Closed-Loop Robot Software}
\author{So Kuroki$^{1}$, Kento Kawaharazuka$^{2}$, and Yujin Tang$^{1}$\\
$^{1}$Sakana AI \qquad $^{2}$The University of Tokyo}
\begin{document}
\maketitle
\thispagestyle{empty}
\pagestyle{empty}
\begin{abstract}
Closed-loop robot policies require observation processing, state management, and situation-dependent branching, making them costly to design and tune manually.
Although coding agents increasingly support control-code generation
and optimization, it remains unclear whether implementations improved
on source tasks also support policy acquisition for new tasks.
We study this question by treating complete closed-loop implementations as reusable execution experience.
To learn each source policy, a coding agent generates policy code from a few successful demonstrations and iteratively improves it using simulation feedback.
The validation-selected implementations are retained in a software archive.
To transfer the acquired software to new tasks, the agent generates and improves policies
using archived implementations, target demonstrations, and execution feedback.
We evaluate policy learning and its relationship to cross-task transfer.
For learning, across four source tasks in RoboCasa, iterative
optimization increases the success rate from 28.3\% with initially generated policy code
to 64.2\% with validation-selected policy code.
We further validate this learning approach on a physical
Unitree G1.
For transfer, across nine additional target tasks in RoboCasa,
success rates are 45.2\% without references, 41.5\% using initially generated policy code from source tasks, and 57.0\% using their validation-selected policy code.
These results show that improving source-task code also improves learning on new tasks.

\end{abstract}
\section{Introduction}
Robots require closed-loop policies that adjust actions to observations
to accomplish diverse tasks reliably.
Implementing these policies as software involves combining object recognition,
motion-phase management, and situation-dependent branches.
Designing and tuning this logic manually for each task is costly.

Coding agents are making the acquisition and revision of such software
increasingly automated.
Code as Policies shows that language models can generate programs using
perception and control APIs to express observation-dependent
policies~\cite{liang2022cap}.
Subsequent work uses execution feedback to rewrite control code,
including closed-loop implementations and multi-file policy
repositories~\cite{kumar2026aor,elmaaroufi2026rho}.
These studies treat code not merely as a fixed procedure,
but as a policy representation updated through experience.
Just as learned weights in neural policies support learning new tasks,
improved control implementations may serve as resources
for acquiring new policies.

However, improving performance on one task does not necessarily yield
an implementation that is useful for another.
Conditional branches and control routines added or tuned during optimization
may become overly specialized to source-task conditions, hindering transfer.
Prior work reuses successful planning or policy code through retrieval and
adaptation~\cite{kagaya2025mtp,tziafas2024lifelong}, or organizes execution experience into
reusable skills, repair knowledge, or functions~\cite{tziafas2024lifelong,lu2026aspire,zhang2026rats}.
In contrast, the value of retaining an entire closed-loop implementation,
including observation processing, state management, and conditional logic,
as a learned artifact of execution experience and using it as a reference
for another task remains unclear.
We therefore ask whether learning better source-task
code also makes it more useful for acquiring policies on target tasks.

\begin{figure*}[tp]
\centering
\includegraphics[width=\textwidth]{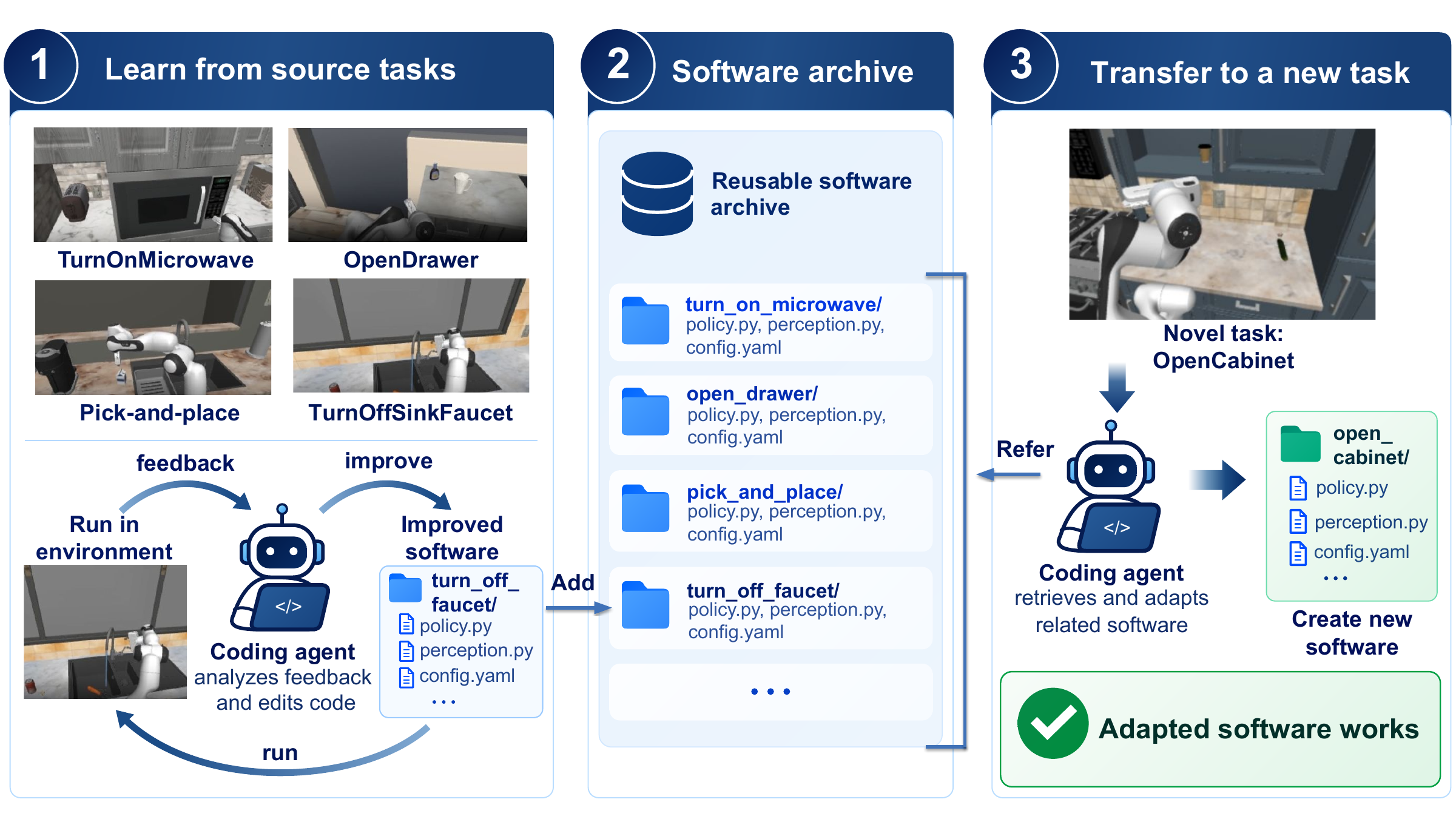}
\caption{Learning and transferring closed-loop robot software.
Source demonstrations initialize a policy that is revised using train traces;
validation scores select an implementation for the archive.
For a target task, the agent references existing implementations,
generates a policy from successful demonstrations,
and adapts the software through improvements using train feedback.}
\label{fig:overview}
\end{figure*}

Our framework learns closed-loop
robot software through execution
and uses the resulting implementations as references for new tasks
(Fig.~\ref{fig:overview}).
A coding agent generates policy code given a few demonstrations,
improves it using simulation feedback, and retains
the validation-selected implementation in an archive.
For a new task, the agent can inspect the archive and select or revise
relevant routines, using existing implementations with situation-dependent
control logic to generate a new policy.

We evaluate policy learning and its relationship to cross-task transfer.
For learning, across four source tasks in RoboCasa~\cite{nasiriany2026robocasa365}, iterative
optimization increases mean success from 28.3\% with initially generated policy code to
64.2\% with validation-selected policy code. We further validate this
learning approach on a physical Unitree G1 across three tasks.
For transfer, across nine additional RoboCasa target tasks,
we compare policy generation and improvement without references, using initially generated policy code from source tasks, and using their validation-selected policy code.
Mean success is 45.2\%, 41.5\%, and 57.0\%, respectively.
Using validation-selected policy code as references outperforms using initially generated policy code in all three runs
on the nine-task average, yielding a mean gain of 15.6 percentage points.
Thus, better-performing code learned on source tasks serves as
a more effective reference for policy acquisition on transfer tasks than
initially generated policy code.

Our contributions are threefold:
\begin{itemize}
\item \textbf{Transfer of learned robot software.}
We study whether improving closed-loop policy code on source tasks makes it more useful for learning new tasks.
\item \textbf{Learning closed-loop policies through execution-guided
optimization.}
We evaluate policy learning in RoboCasa and validate the approach on a physical Unitree G1.
\item \textbf{Source-task learning versus transferability.}
Across nine target tasks, we show that validation-selected code improves transfer performance on average over initially generated code and no references.
\end{itemize}

\section{Related Work}
\subsection{Code as a Learnable Policy Representation}
Representing policies and plans as code turns robot behavior acquisition
into program generation and revision.
Code as Policies and related approaches connect perception with planning
and control through generated programs, enabling observation-dependent
computation and action
selection~\cite{liang2022cap,huang2023voxposer,huang2024rekep}.
Beyond language specifications, demonstration-conditioned approaches
infer procedures or conditions to acquire programmatic
policies~\cite{wang2023demo2code,xin2024plunder,murray2025showtell}.
Our work shares this view of code as a policy representation acquired
from task information and updated through execution experience.
We generate an initial policy from a few demonstrations and optimize
the resulting implementation as a resource for acquiring policies for new tasks.

\subsection{Execution-Guided Code Optimization}
Updating generated programs through experience allows policies
to address situations their initial implementations cannot handle.
Program-policy repair from additional demonstrations can modify
action-selection predicates as well as
parameters~\cite{holtz2021idips}.
More recent coding-agent frameworks use observations, rewards,
and failure records to revise controllers and multi-file policy
implementations between trials~\cite{kumar2026aor,elmaaroufi2026rho}.
The observation, control-interface, and feedback choices supporting
embodied coding agents have also been studied~\cite{fu2026capx}.
We share the use of execution feedback to revise closed-loop implementations
and apply it to source policy acquisition.
However, conditional branches and control routines may overfit the source task. We therefore ask whether improving source-task code also makes it more useful as a reference for learning new tasks.

\subsection{Transferring Code and Repair Knowledge}
Reusing experience across tasks requires choosing what to retain.
Skill and repair-memory approaches preserve applicability conditions,
code examples, or executable functions to support subsequent code generation
and execution~\cite{lu2026aspire,zhang2026rats,chen2026architect}.
Memory Transfer Planning (MTP) retrieves and adapts successful planner-level code;
its stored and retrieved units do not include the Composer
or low-level language-model programs~\cite{kagaya2025mtp}.
Rather than requiring skill extraction or retaining only planner code,
we retain complete closed-loop implementations containing observation processing,
state management, and conditional logic as learned artifacts of execution experience.
For a new task, the coding agent uses these implementations as references
and selects or revises the necessary routines.

\section{Learning and Transferring Closed-Loop Robot Software}
\label{sec:method}
Our method retains closed-loop control implementations improved through execution
and uses them as references when acquiring policies for new tasks
(Fig.~\ref{fig:overview}).

\subsection{Task Definition and Policy Interface}
\label{sec:method-representation}
For a task $q$, we seek a policy $P$ that maps observations to actions
and satisfies the task's success condition within a fixed horizon.
Let $x_t$ be the simulator state and $\omega$ the observation configuration.
A closed-loop policy implements
\begin{equation}
\begin{aligned}
o_t &= \mathcal{O}_{\omega}(x_t),\\
(a_t,h_{t+1}) &= P(o_t,h_t;q).
\end{aligned}
\label{eq:policy}
\end{equation}
Here, $o_t$ contains the permitted observations, $a_t$ is a controller command,
and $h_t$ is internal state, such as the current manipulation phase.
The internal state is reset at the beginning of each episode.
After each action, the policy receives a new observation and continues control.
The observation configuration $\omega$ is fixed during code optimization.

The coding agent $\mathcal{A}_B$ receives a task instruction,
successful demonstrations $\mathcal{D}_q$, and the observation--action
interface specification $\mathcal{I}_{\omega}$.
The budget $B$ constrains code generation.
Using these inputs and the additional information permitted at each stage,
the agent outputs an executable policy package $P$ containing observation
processing, conditional logic, and control code.
Learning proceeds by repeatedly executing and refining this code and its configuration using execution feedback, while the coding model weights remain fixed.
The resulting policy executes without calling the coding model.

For each source task, cases with different initial states are partitioned
into disjoint train, validation, and test sets before optimization.
Train provides execution experience for code revision;
validation supports candidate comparison and selection;
test evaluates the selected, frozen policy.
We denote the success rate of $P$ on cases $\mathcal{C}$
by $\widehat{J}_{\mathcal{C}}(P)$.
Source test results are not used for revision or reselection.

\subsection{
Learning and Retaining Implementations}
\label{sec:method-optimization}
We learn a policy for each source task $s$ by generating and iteratively optimizing its code, then retain the validation-selected policy code as a reference for new tasks.
The initial policy $P_s^{(0)}$ is generated from successful demonstrations
$\mathcal{D}_s$, the task instruction, and $\mathcal{I}_{\omega}$.
The objective is not demonstration replay, but a policy that responds
to new observations according to Eq.~\eqref{eq:policy}.

Each candidate is executed in simulation.
The agent receives train observations, actions, task outcomes, and execution errors.
From validation, it receives only aggregate success and evaluation counts,
not individual images or trajectories.
Using the candidate code and this feedback, it revises observation processing,
control parameters, conditions, and state transitions.
Previously evaluated candidates are ranked by train success, with ties favoring earlier generations, to select the starting point for each revision.

After $K$ generations, including
initial code generation,
an implementation is selected by validation success:
\begin{equation}
P_s^\star \in
\underset{P\in\{P_s^{(0)},\ldots,P_s^{(K-1)}\}}
{\arg\max}\ \widehat{J}_{\mathcal{C}_s^{\mathrm{val}}}(P).
\label{eq:selection}
\end{equation}
Here, $\mathcal{C}_s^{\mathrm{val}}$ is the validation case set for source task $s$.
The selected implementation is frozen and evaluated on test.
Implementations from source tasks $\mathcal{S}$ form the archive
$\mathcal{M}=\{P_s^\star\mid s\in\mathcal{S}\}$.
We retain the code and assets needed for execution,
without requiring context-dependent logic to be distilled into reusable skills.

\subsection{Transfer Through Archive Access}
\label{sec:method-transfer}
For a new task $q$, the agent can use archived implementations as references.
In addition to the target instruction, demonstrations, and interface,
it receives read-only access to $\mathcal{M}$ and generates an initial policy:
\begin{equation}
\widetilde{P}_q^{(0)} =
\mathcal{A}_B(q,\mathcal{D}_q,\mathcal{I}_{\omega},\mathcal{M}).
\label{eq:adapt}
\end{equation}
No source--target mapping is imposed.
The agent chooses which implementations to inspect and which parts
to reuse, combine, or modify.

The initial policy $\widetilde{P}_q^{(0)}$ is executed on target train cases.
Observation--action trajectories and outcomes are returned to the agent
for improvement.
Using this feedback, the agent revises $\widetilde{P}_q^{(0)}$ to obtain an improved policy $\widetilde{P}_q$.
The final policy $\widetilde{P}_q$ is frozen and evaluated on test.

\begin{figure*}[!t]
\centering
\includegraphics[width=\textwidth]{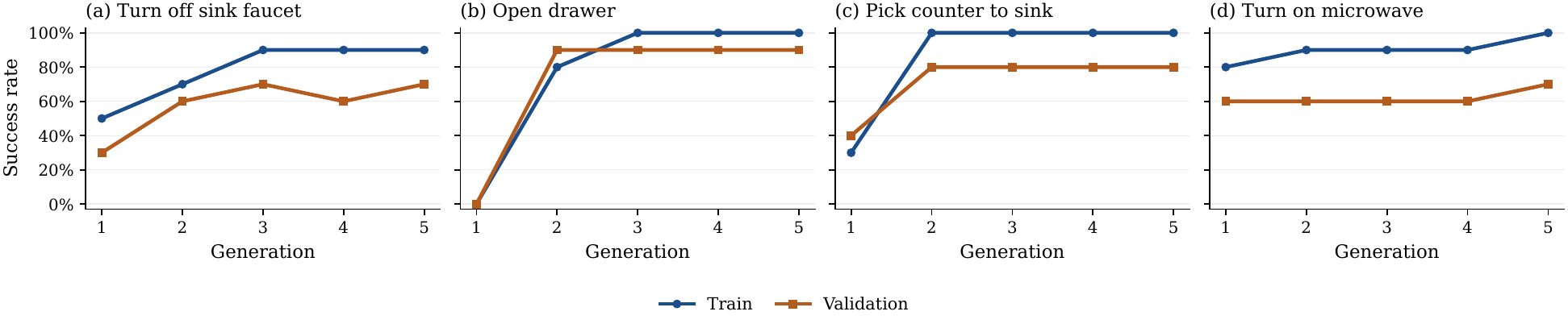}
\caption{Train and validation success across effective generations
for the four source tasks.
Each split contains ten cases.
Candidates that fail before a simulator step are excluded;
when a raw generation has retries, the last evaluated candidate is shown.}
\label{fig:train-validation}
\end{figure*}

\section{Experiments}
\label{sec:experiments}
We evaluate whether execution experience yields effective closed-loop software
and whether the acquired implementations help construct policies for new tasks.
Table~\ref{tab:source} and Fig.~\ref{fig:train-validation}
evaluate source-policy learning through iterative code optimization.
Next, Table~\ref{tab:observations} analyzes which observation information
supports acquiring higher-performing policies through optimization.
Finally, Table~\ref{tab:transfer} evaluates whether implementations acquired
on source tasks also help acquire policies for new tasks.

\begin{table*}[!t]
\centering
\caption{Source policy success on the same thirty test cases per task.
Cells report success rates (\%) with counts out of thirty; Mean is the four-task macro-average.
Panel A compares code policies, an online policy, and same-demonstration imitation learning.
Panel B provides reference comparisons with public checkpoints that differ in pretraining and observations, with possible overlap between their training data and the demonstrations underlying our test cases.
Language instructions are supplied where supported.
}
\label{tab:source}
\footnotesize
\setlength{\tabcolsep}{3.5pt}
\begin{tabular}{@{}lccrrrrr@{}}
\toprule
Method & \shortstack{Demos\\per task} & Runtime input &
Faucet & Drawer & Pick$\to$Sink & Microwave & Mean $\uparrow$\\
\midrule
\multicolumn{8}{@{}l}{\textit{A. Code policies, online control, and same-demonstration reference}}\\
Validation-selected policy code (ours) & 10 & Full &
56.7~(17/30) & 70.0~(21/30) & 76.7~(23/30) & 53.3~(16/30) &
\textbf{64.2}\\
Initially generated policy code & 10 & Full &
33.3~(10/30) & 3.3~(1/30) & 26.7~(8/30) & 50.0~(15/30) &
28.3\\
GPT-6 Astra online & 10 & Full &
70.0~(21/30) & 50.0~(15/30) & 60.0~(18/30) & 26.7~(8/30) &
51.7\\
BC-Transformer-10 & 10 & 3 RGB + proprio &
23.3~(7/30) & 6.7~(2/30) & 3.3~(1/30) & 3.3~(1/30) &
9.2\\
\midrule
\multicolumn{8}{@{}l}{\textit{B. Public multi-task checkpoints; reference comparisons}}\\
GR00T N1.5 & 100 & 3 RGB + state &
66.7~(20/30) & 56.7~(17/30) & 90.0~(27/30) & 36.7~(11/30) &
62.5\\
$\pi_{0.5}$ / Human300 & 100 & 3 RGB + state &
50.0~(15/30) & 80.0~(24/30) & 83.3~(25/30) & 20.0~(6/30) &
58.3\\
$\pi_0$ / Human300 & 100 & 2 RGB + state &
46.7~(14/30) & 30.0~(9/30) & 66.7~(20/30) & 6.7~(2/30) &
37.5\\
Diffusion Policy / Human300 & 100 & 3 RGB + proprio &
26.7~(8/30) & 13.3~(4/30) & 30.0~(9/30) & 10.0~(3/30) &
20.0\\
\bottomrule
\end{tabular}
\end{table*}

\subsection{Environment and Evaluation Protocol}
\label{sec:experimental-setup}
We use RoboCasa with a PandaOmron embodiment
and a fixed task registry~\cite{nasiriany2024robocasa,nasiriany2026robocasa365}.
The four source tasks are turning off the sink faucet,
opening a drawer, moving an object from the counter to the sink,
and turning on the microwave.
They cover articulated-object manipulation, object transport,
and contact-dependent appliance operation.

\noindent\textbf{Input information.}
The Full configuration supplies RGB, Metric 3D, ground-truth (GT) geometry information (GT-Geom),
binary contact, and robot state at each policy step.
RGB images come from three cameras: left, right, and wrist-mounted.
Metric 3D comprises metric depth and world-frame point clouds
reconstructed using camera calibration, with each point aligned to an image pixel.
GT-Geom consists of simulator-provided per-pixel geometry IDs
and name/class metadata indicating the object or part associated with each ID.
The mask contains no depth itself, but aligned 3D points
associate an object's image region with its spatial location.
Binary contact signals indicate external-object contact
for the robot, grippers, and fingers, without contact forces or object IDs.
Robot state includes joint positions and velocities,
world-frame end-effector poses, gripper and other part joint positions,
and controller-frame poses.
The agent also receives the task instruction and action interface,
with camera intrinsics and camera-to-world transforms available.

\begin{table*}[tp]
\centering
\caption{Observation ablation with five train cases, five validation cases,
  and ten test cases per task and condition.
  Optimization runs for three generations evaluated in simulation,
  including the initial policy.
  Task columns report successes out of ten; Mean is the four-task macro-average
  success rate.
  $\Delta$ Full is the condition mean minus the Full mean, in percentage points.
  }
\label{tab:observations}
\footnotesize
\setlength{\tabcolsep}{3.5pt}
\begin{tabular}{@{}lccccrrrrrr@{}}
\toprule
Observation & RGB & 3D & GT-Geom & Contact & Faucet & Drawer & Pick$\to$Sink &
Microwave & \shortstack{Mean (\%)\\$\uparrow$} & \shortstack{$\Delta$ Full\\(pp)}\\
\midrule
RGB only & $\checkmark$ &  &  &  &
2/10 & 4/10 & 2/10 & 1/10 & 22.5 & $\textbf{-42.5}$\\
Without Metric 3D & $\checkmark$ &  & $\checkmark$ & $\checkmark$ &
8/10 & {10/10} & {7/10} & 2/10 & 67.5 & $+2.5$\\
Without GT-Geom & $\checkmark$ & $\checkmark$ &  & $\checkmark$ &
5/10 & 8/10 & 1/10 & 0/10 & 35.0 & $\textbf{-30.0}$\\
Without contact & $\checkmark$ & $\checkmark$ & $\checkmark$ &  &
8/10 & 9/10 & 6/10 & {5/10} & {70.0} & $+5.0$\\
\textbf{Full} & $\checkmark$ & $\checkmark$ & $\checkmark$ & $\checkmark$ &
{10/10} & 9/10 & 6/10 & 1/10 & 65.0 & $0.0$\\
\bottomrule
\end{tabular}
\end{table*}

\noindent\textbf{Observation conditions.}
The code and online policies in Table~\ref{tab:source},
and the transfer conditions in Table~\ref{tab:transfer}, use Full.
Table~\ref{tab:observations} compares Full with three conditions
removing Metric 3D, GT-Geom, or contact individually,
and an RGB-only condition removing all three.
Removing Metric 3D removes both depth and derived points;
removing GT-Geom removes both masks and labels.
Task instructions, calibration, robot state, action history,
and the action interface are shared across conditions.
Code is generated and optimized anew under each condition.

For Table~\ref{tab:source}, source optimization uses ten successful demonstrations per task,
with ten train cases corresponding one-to-one to the demonstrations
and sharing their environments and initial states.
Ten validation cases and thirty test cases, disjoint from train and each other,
are used for policy selection and final evaluation, respectively.
For each task, initially generated policy code denotes the first executable candidate before execution-guided improvement and constitutes generation one.
We alternate evaluation and improvement using execution feedback
until generation five has been evaluated.
For Table~\ref{tab:observations}, the observation ablation uses five train, five validation, and ten test cases,
running the improvement loop through generation three for each task and condition.

Matched comparisons share case lists, embodiment, task horizons, success predicates, and simulator settings.
The evaluation horizons are 300 steps for the faucet, 750 for the drawer,
600 for pick-and-place into the sink, and 450 for the microwave.
The coding model is \texttt{gpt-6-astra} with high reasoning effort.

\subsection{
Closed-Loop Policy Learning and Baseline Comparisons}
\noindent\textbf{Effect of iterative optimization.}
Iterative optimization improves on the initially generated policy code
(Table~\ref{tab:source}).
Using the same ten demonstrations and Full observations,
execution-guided revision followed by validation selection increases
mean test success from 28.3\% (34/120) to 64.2\% (77/120),
a gain of 35.8 percentage points.
All four tasks improve.
Train and validation success show the same upward trend
across tasks in Fig.~\ref{fig:train-validation};
improvement is not confined to train cases.
See Sec.~\ref{sec:real_robot} for physical-robot validation.

\noindent\textbf{Direct use of the coding agent.}
Our method exceeds the online use of the same coding agent,
which achieves 51.7\% mean success, by 12.5 percentage points.
The online condition receives the same ten demonstrations and Full observations,
and repeatedly generates short action sequences, executes them,
and obtains updated observations.
It requires 52.6 model calls per episode on average,
whereas our frozen code executes without calls to the coding model.

\noindent\textbf{Few-shot imitation learning.}
Our method exceeds the 9.2\% mean success of BC-Transformer-10,
trained separately per task on the same ten successful
demonstrations~\cite{nasiriany2024robocasa}.
Following each method's input specification, BC-Transformer receives
three RGB views and robot state, while our method receives Full observations.
Our method also uses execution experience during optimization,
so we present this as a reference comparison with different input and training conditions.

\noindent\textbf{Public multi-task policies.}
Our method optimizes code separately for each task,
whereas the public policies share weights trained across multiple
RoboCasa tasks~\cite{nasiriany2026robocasa365}.
The public policies receive RGB and robot state.
We report their results as reference comparisons given the different training and input conditions.
Our mean success numerically exceeds the VLA foundation models
GR00T N1.5 at 62.5\%~\cite{bjorck2025groot,nvidia2025groot15},
$\pi_{0.5}$ at 58.3\%~\cite{physicalintelligence2025pi05},
and $\pi_0$ at 37.5\%~\cite{black2024pi0}.
We use the public Human300 Diffusion Policy without additional training, which reaches 20.0\%~\cite{nasiriany2026robocasa365, chi2023diffusion}.

\subsection{
Inputs Supporting Effective Policy Learning}
To complement the preceding policy-learning evaluation, we analyze which observation information supports acquiring higher-performing policies.
Table~\ref{tab:observations} evaluates policy learning under the five observation conditions defined in Sec.~\ref{sec:experimental-setup}.

Mean success drops from 65.0\% with Full to 22.5\% with RGB only.
Among single-group removals, removing GT-Geom causes the largest drop,
to 35.0\%, or 30.0 percentage points below Full.
This suggests that the additional observations, particularly GT-Geom,
support the acquisition of effective control code in this setting.
The individual contributions of pixel segmentation and labels,
and generalization to estimated real-world segmentation, remain unevaluated.

Without Metric 3D and without contact, mean success is 67.5\% and 70.0\%,
respectively, slightly above Full.
With one optimization run per task and condition, these differences do not
establish that removing information is generally beneficial.

\begin{figure*}[!t]
\centering
\includegraphics[width=\textwidth]{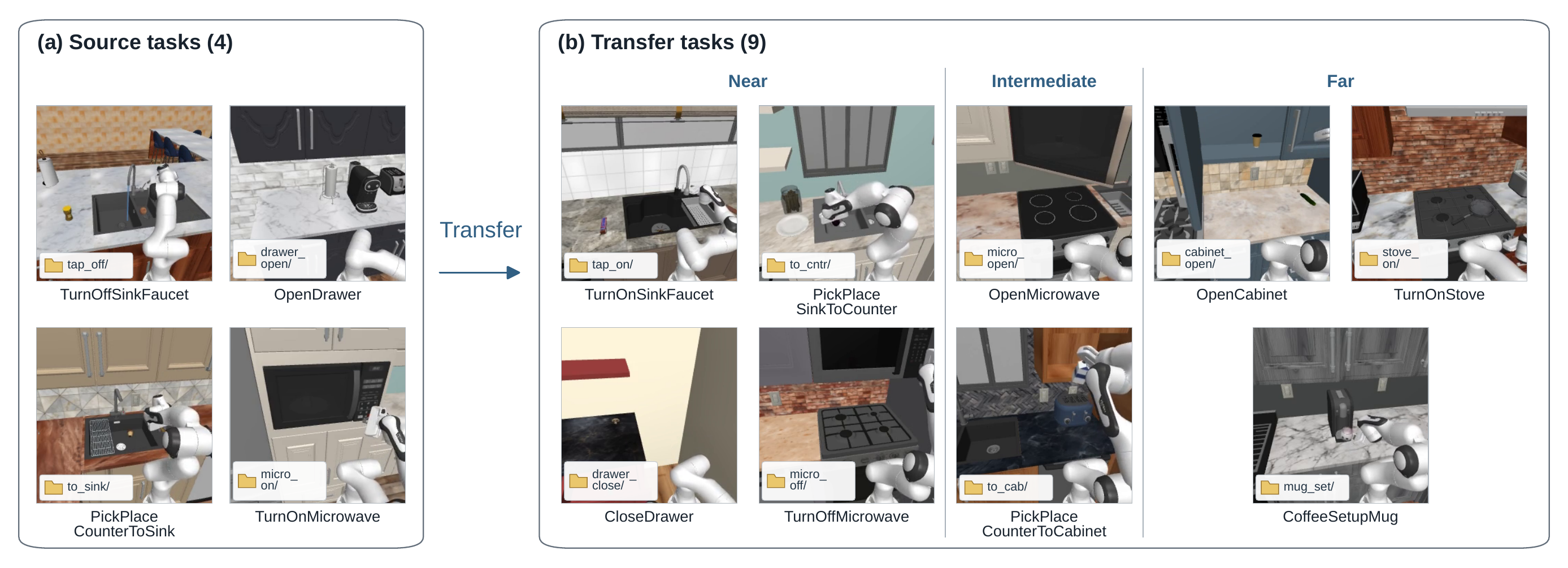}
\caption{Source and target tasks for software transfer.
Four source tasks are shown on the left and nine target tasks on the right.
Targets are grouped as Near, Intermediate, and Far by task-structure similarity.}
\label{fig:task-transfer}
\end{figure*}

\begin{table*}[tp]
\centering
\caption{
Transfer with initially generated and validation-selected policy code over three independent runs.
Initial4 exposes the initially generated policy code from the four source tasks. Best4 exposes their validation-selected policy code.
In each run, policies are generated from five successful demonstrations,
improved using five train cases, and evaluated on ten shared test cases per task.
Cells report mean success rate $\pm$ sample standard deviation (\%)
across the three runs, with total successes out of thirty.
The overall row summarizes the nine-task macro-average in each run.
}
\label{tab:transfer}
\footnotesize
\setlength{\tabcolsep}{3pt}
\begin{tabular}{@{}llrrr@{}}
\toprule
Transfer & Target task & No reference $\uparrow$ & Initial4 $\uparrow$ & Best4 $\uparrow$\\
\midrule
Near & TurnOnSinkFaucet &
90.0 $\pm$ 17.3~(27/30) & 76.7 $\pm$ 32.1~(23/30) & 90.0 $\pm$ 10.0~(27/30)\\
Near & PickPlaceSinkToCounter &
60.0 $\pm$ 52.0~(18/30) & 0.0 $\pm$ 0.0~(0/30) & 70.0 $\pm$ 20.0~(21/30)\\
Near & CloseDrawer &
70.0 $\pm$ 52.0~(21/30) & 86.7 $\pm$ 5.8~(26/30) & 93.3 $\pm$ 11.5~(28/30)\\
Near & TurnOffMicrowave &
3.3 $\pm$ 5.8~(1/30) & 76.7 $\pm$ 15.3~(23/30) & 60.0 $\pm$ 10.0~(18/30)\\
\midrule
Intermediate & OpenMicrowave &
80.0 $\pm$ 17.3~(24/30) & 23.3 $\pm$ 40.4~(7/30) & 83.3 $\pm$ 20.8~(25/30)\\
Intermediate & PickPlaceCounterToCabinet &
63.3 $\pm$ 23.1~(19/30) & 53.3 $\pm$ 23.1~(16/30) & 56.7 $\pm$ 32.1~(17/30)\\
\midrule
Far & OpenCabinet &
23.3 $\pm$ 20.8~(7/30) & 40.0 $\pm$ 10.0~(12/30) & 40.0 $\pm$ 10.0~(12/30)\\
Far & TurnOnStove &
16.7 $\pm$ 15.3~(5/30) & 6.7 $\pm$ 11.5~(2/30) & 20.0 $\pm$ 20.0~(6/30)\\
Far & CoffeeSetupMug &
0.0 $\pm$ 0.0~(0/30) & 10.0 $\pm$ 17.3~(3/30) & 0.0 $\pm$ 0.0~(0/30)\\
\midrule
\multicolumn{2}{@{}l}{Mean (9 tasks, 3 runs)} &
45.2 $\pm$ 10.3~(122/270) & 41.5 $\pm$ 10.5~(112/270) & \textbf{57.0 $\pm$ 7.8}~(154/270)\\
\bottomrule
\end{tabular}
\end{table*}

\subsection{
Cross-Task Transfer of Learned Software}
\label{sec:transfer-experiments}
Figure~\ref{fig:task-transfer} and Table~\ref{tab:transfer}
evaluate the effect of using four implementations acquired on the source tasks
in Table~\ref{tab:source} as references for generating and improving policies
on nine new tasks.
Task-structure similarity defines four Near, two Intermediate,
and three Far targets.
Near transfers change operations or directions within similar mechanisms;
Intermediate transfers include different operations on the same appliance
or changes in placement context;
Far transfers involve different devices or operations.

\noindent\textbf{Comparison conditions.}
No reference provides no existing policy.
Initial4 provides the initially generated policy code from the four source tasks,
identical to the initial-code baseline in Table~\ref{tab:source}.
Best4 provides the validation-selected policy code from the source tasks.
Both reference conditions expose four implementations as read-only resources.
No source--target pairing is specified:
the agent decides which implementations to inspect and which parts to reuse.
All conditions start from the same minimal policy scaffold.

For each task and condition, the agent generates a policy from five shared
successful demonstrations and improves it once using five train rollouts.
If the initial train execution encounters a code error,
the repaired code is rerun on the same five train cases,
and those trajectories support one further improvement.
The final code is frozen and evaluated on ten shared test cases,
without target validation selection or manual policy edits.
The model, Full observations, cases, and per-session limits
(30 minutes and 80 API calls) are shared.
Error-triggered execution and improvement add computation,
so total compute differs across conditions.

We repeat target policy generation, improvement, and test evaluation
independently three times for each task and condition, using the same source
archives, demonstrations, and case sets.
Table~\ref{tab:transfer} reports means and sample standard deviations
across the three runs, with thirty test evaluations per task and condition.
For the overall result, we first compute the nine-task macro-average
within each run and then summarize these three averages.

\noindent\textbf{
Source-policy learning and transfer.}
Best4 achieves $57.0\pm7.8\%$ mean success,
compared with $45.2\pm10.3\%$ without references and $41.5\pm10.5\%$ with Initial4.
Best4 outperforms Initial4 on the nine-task average in all three runs,
with a mean improvement of 15.6 percentage points.

On the per-task three-run means, Best4 exceeds Initial4
on six tasks, ties on OpenCabinet, and is lower on TurnOffMicrowave
and CoffeeSetupMug.
The largest gains occur on PickPlaceSinkToCounter (70.0 percentage points)
and OpenMicrowave (60.0 percentage points).
Initial4 does not improve the overall mean over no reference,
showing that access to existing code alone is not sufficient to improve performance
in this evaluation. Learning higher-performing policy code improves its value as a reference.

The remaining task-level reversals may reflect
source-specific specialization, reference selection, or
policy-generation variability.
Static analysis of the reference implementations' Python code shows that
the total number of \texttt{if}/\texttt{elif} statements, excluding test code,
increases from 622 in Initial4 to 741 in Best4 (19.1\%).
All four implementations show an increase.
Although this increase alone does not establish specialization or adaptation difficulty,
situation-dependent branches and control routines can support source-task success
while requiring revision for new tasks.

\begingroup

\section{Physical-Robot Validation}
\label{sec:real_robot}
To examine whether
policy learning through coding-agent-driven optimization is applicable
to real-world settings, we deploy simulation-optimized code on a physical
Unitree G1 for drawer closing, refrigerator-door closing, and navigation
to a refrigerator or cabinet followed by stopping.
This validates the learning component on a physical humanoid, without evaluating archive-based cross-task transfer.

\begin{figure}[!ht]
\color{black}
\centering
\begin{minipage}[t]{0.32\linewidth}
\centering\includegraphics[width=\linewidth]{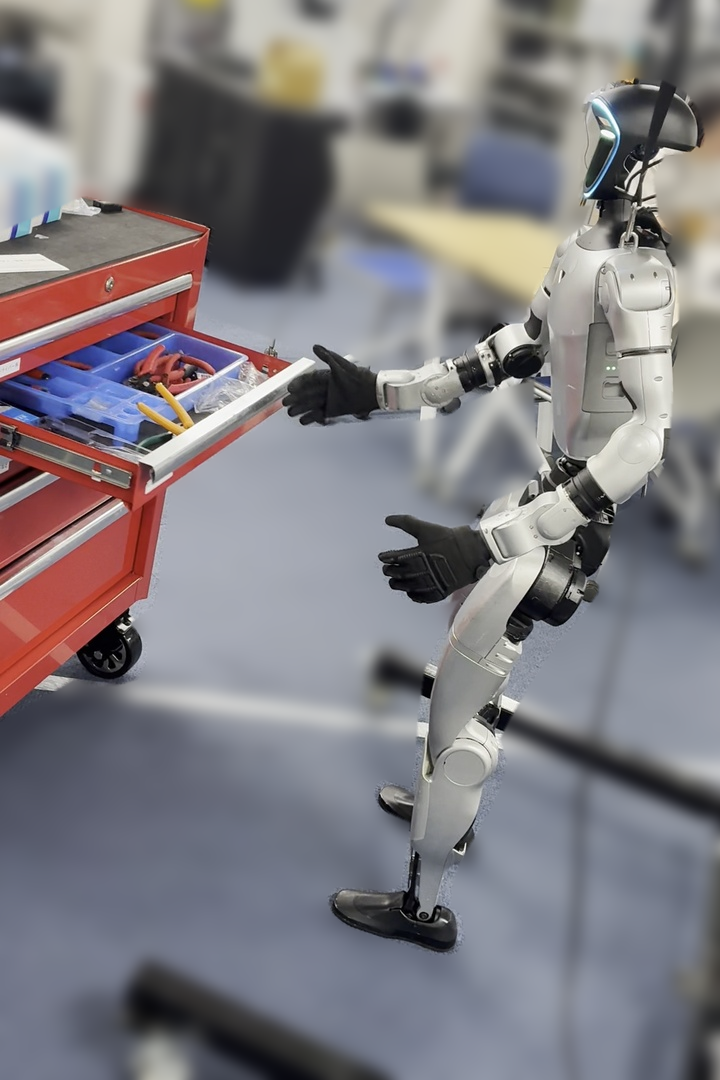}\\[-1pt]
{\scriptsize\color{black}(a) Close drawer}
\end{minipage}\hfill
\begin{minipage}[t]{0.32\linewidth}
\centering\includegraphics[width=\linewidth]{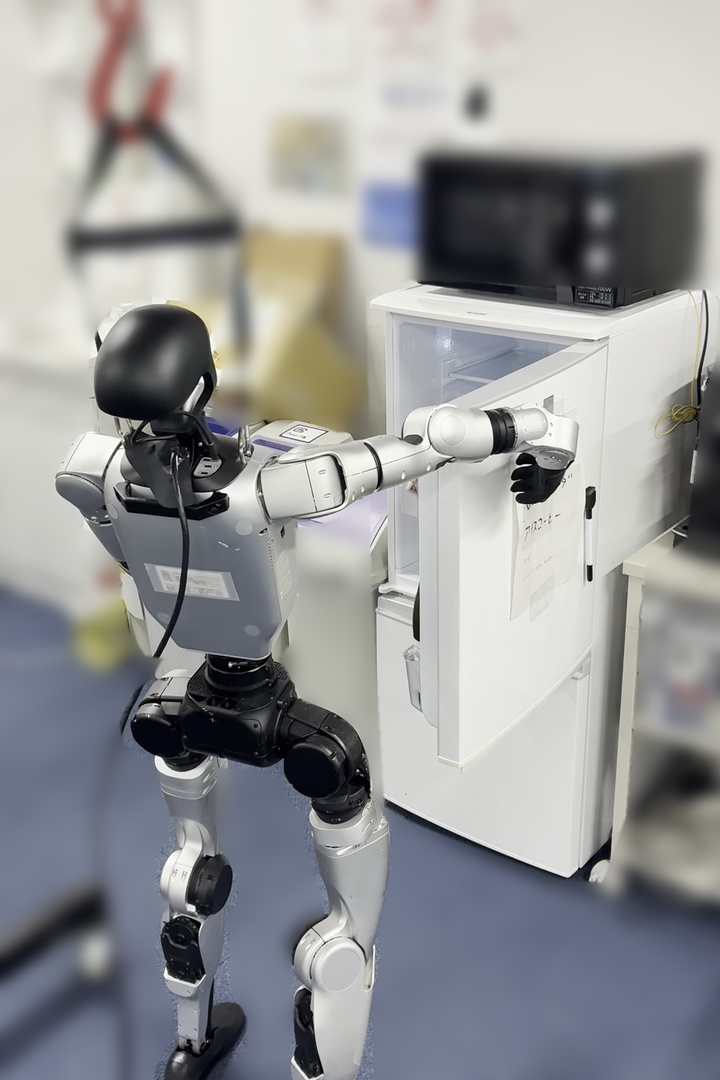}\\[-1pt]
{\scriptsize\color{black}(b) Close refrigerator}
\end{minipage}\hfill
\begin{minipage}[t]{0.32\linewidth}
\centering\includegraphics[width=\linewidth]{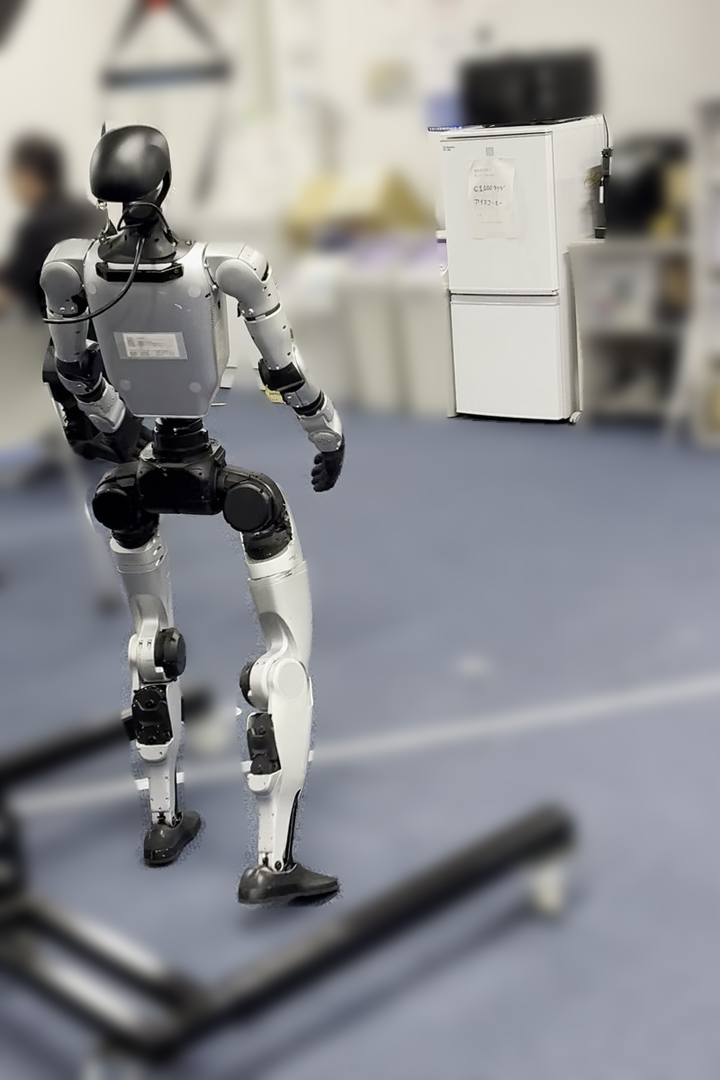}\\[-1pt]
{\scriptsize\color{black}(c) Navigation}
\end{minipage}
\caption{G1 hardware sessions (navigation shown toward a
refrigerator). Backgrounds are blurred for privacy.}
\label{fig:g1_real_photos}
\end{figure}

\noindent\textbf{Setup.}
The robot uses 29 actuated body joints and fixed five-finger hands.
The generated policy code runs at 50\,Hz above SONIC, specifying locomotion
and head/hand pose targets while SONIC produces the corresponding whole-body
motion~\cite{luo2026sonic}.
The policy inputs are designed to be obtainable on the physical robot.
The policy receives robot state and depth-derived point
clouds. JEV-1.14, a decision model from TypeSafe~\cite{typesafe2026jev}, identifies target regions in images. Our perception and policy code uses these regions with depth-derived point clouds to estimate target positions and geometry.
The policy does not receive simulator ground-truth object poses or contact labels.
The camera is configured with a
$20^\circ$ downward mounting angle.

\noindent\textbf{Optimization in simulation.}
For each humanoid task, we optimize policy code in RoboCasa on ten train cases without demonstrations.
Claude Opus 5.5 generates and revises code using execution feedback.
We run up to eight generations in simulation, evaluate each on the same train cases, and select the candidate with the highest train success.
 The loop stops early when a candidate succeeds on all ten
cases. Figure~\ref{fig:g1_real_curve} shows that train performance improves
overall across generations despite intermediate variation.
The selected candidates are from generations~4, 7, and~7 for drawer, refrigerator, and navigation, respectively.

\begin{figure}[!ht]
\color{black}
\centering
\includegraphics[width=0.80\linewidth]{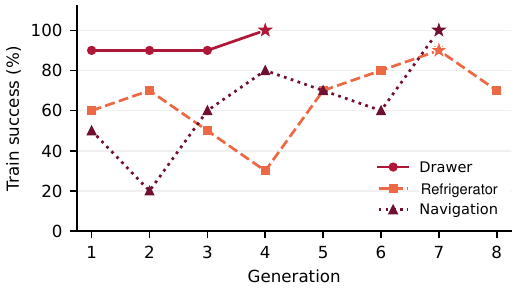}
\caption{Simulation train success (ten cases per generation).
Stars mark deployed candidates. Drawer and navigation end at early
stopping.}
\label{fig:g1_real_curve}
\end{figure}

\noindent\textbf{Physical results.}
Success requires fully closing the drawer or refrigerator door, or stopping in front of the target for navigation.
Before deployment, we make small task-specific adjustments to runtime parameters, including target distances, to match the physical setup. Table~\ref{tab:g1_real_results} shows that the selected policies largely transfer to previously unseen physical objects without real-to-simulation reconstruction, despite differences in object geometry and placement.
 Differences in refrigerator distance and height, for example,
contribute to failures that remain in the physical evaluation.

\begin{table}[!ht]

\centering
\caption{Physical evaluation of the selected policies.}
\label{tab:g1_real_results}
\small
\begin{tabular}{lr}
\toprule
Task & Successes / trials \\
\midrule
Close drawer & 4/5 \\
Close refrigerator & 3/5 \\
Navigation & 4/5 \\
\bottomrule
\end{tabular}
\end{table}
\par
\endgroup

\section{Limitations and Future Work}
\label{sec:limitations}
First, while validation-selected source policies improve transfer
performance on average, they are not the most effective references for every
target task, motivating further investigation into when source-task
improvements lead to better transfer. Second, our physical evaluation combines generated high-level policy code with SONIC's low-level whole-body controller and does not isolate their respective contributions. Third, while the learned policies
successfully perform three tasks on a physical humanoid through a whole-body
controller, more extensive evaluation is needed to assess their robustness
across a wider range of tasks, objects, and physical conditions.

\section{Conclusion}

We investigated whether learning better-performing closed-loop policies through execution-guided code optimization on source tasks also makes their implementations more useful for acquiring policies on new tasks.
For learning, across four source tasks in RoboCasa,
iterative optimization increases mean success from 28.3\% to 64.2\%.
The observation ablation identifies GT-Geom as an important input for acquiring effective code.
We further validate this learning approach on a physical
Unitree G1 across three tasks.
For transfer, across nine additional RoboCasa target tasks
and three independent runs, Best4 achieves
57.0\% mean success, compared with 41.5\% for Initial4 and 45.2\%
without references.
These findings support
retaining validation-selected policy code
as a resource for subsequent policy acquisition.

\section*{Acknowledgments}
This work was partially supported by JST Moonshot R\&D Program Japan Grant Number JPMJMS263D.

\bibliographystyle{IEEEtran}
\bibliography{references_updated_arXiv}
\end{document}